\documentclass[11pt]{article}

\usepackage[letterpaper,margin=0.8in]{geometry}
\usepackage{amsmath,amssymb,amsfonts}
\usepackage{graphicx}
\usepackage{booktabs}
\usepackage{multirow}
\usepackage{array}
\usepackage{tabularx}
\usepackage{enumitem}
\usepackage{microtype}
\usepackage{hyperref}
\usepackage{xcolor}
\usepackage{caption}
\usepackage{subcaption}
\usepackage{placeins}
\usepackage{tikz}
\usepackage{float}
\usetikzlibrary{arrows.meta,positioning,calc}

\hypersetup{
    colorlinks=true,
    linkcolor=blue,
    citecolor=blue,
    urlcolor=blue
}

\newcommand{\R}{\mathbb{R}}

\title{\textbf{Confusion-Adaptive Margin Learning for Handwritten Letter Recognition:
A Controlled Study on EMNIST Letters}}

\author{
Azam Nouri\\
Department of Science, Technology \& Mathematics, Lincoln University
}
\date{}

\begin{document}

\maketitle

\begin{abstract}

Handwritten character recognition systems can achieve high overall accuracy,
yet their remaining errors are often concentrated among a small number of
visually similar classes. This study extends our earlier work on handwriting
recognition using planar curvature and gradient orientation
\cite{NouriCurv2025, nouri2025ijca}, in which structured confusion among similar
letters remained visible even after effective geometric representation. We
investigate whether such residual ambiguities can instead be addressed through
the learning objective.

We introduce \textbf{Confusion-Adaptive Margin Learning (CAML)}, a training
strategy that constructs a sparse weighted confusion graph from a separate
discovery partition held out from pilot training. The strongest empirically
observed rival class pairs are then assigned pair-specific cosine-separation
constraints during final training. CAML modifies only the training objective;
it introduces no additional classifier, routing rule, or inference-time
parameter.

CAML is evaluated on EMNIST Letters against a strictly matched cross-entropy
baseline using the same CNN architecture, final training data, deterministic
augmentation schedule, optimizer, and checkpoint-selection procedure. Eight
confusion pairs are discovered: I/L, G/Q, U/V, R/V, D/O, C/E, I/J, and H/N.
On the clean official test set, accuracy changes from \(95.6346\%\) for the
baseline to \(95.6731\%\) for CAML, while macro-F1 changes from \(0.956326\)
to \(0.956724\). Across ten predefined corrupted conditions, mean accuracy
increases from approximately \(95.3812\%\) to \(95.4303\%\). Within the eight
targeted confusion pairs, symmetric clean-test errors decrease from 614 to 605,
a \(1.47\%\) reduction.

The global differences are small and not statistically reliable. On clean
data, CAML fixes 124 baseline errors while introducing 116 new errors
(\(p=0.6515\), exact McNemar test), and none of the corruption-specific
differences remains significant after Holm correction. The results therefore
do not establish CAML as a globally superior recognizer. Rather, they show
that a leakage-controlled, pair-specific margin objective can alter targeted
error structure while producing only limited aggregate benefit when the
underlying classifier is already strong.

\end{abstract}


\section{Introduction}
\label{sec:introduction}

Handwritten character recognition remains a useful setting for studying
fine-grained visual classification because it combines substantial
within-class variability with strong similarity between particular classes.
Variations in stroke thickness, slant, local deformation, writing style, and
incomplete or exaggerated strokes can make instances of the same character
appear quite different. At the same time, different letters may share most of
their global structure and differ only through a small opening, terminal
stroke, loop, junction, or tail. Consequently, high aggregate recognition
accuracy can coexist with persistent errors concentrated around a relatively
small number of class boundaries.

Modern neural approaches, particularly convolutional networks, have greatly
improved handwritten-image recognition by learning spatial features directly
from pixels \cite{lecun1998gradient}. However, standard classification
accuracy does not reveal whether residual errors are distributed broadly or
concentrated among a small number of visually similar classes. For a strong
classifier, this distinction becomes increasingly important because further
performance gains may depend less on improving all classes uniformly and more
on understanding the specific relationships that remain difficult.

The present study builds directly on our earlier work on handwritten-character
recognition using planar curvature and gradient orientation
\cite{nouri2025ijca,NouriCurv2025}. That work investigated explicit geometric
representations designed to capture local shape information in handwritten
characters. Its error analysis showed that residual mistakes were strongly
structured rather than uniformly distributed across the label space, with
visually similar letters accounting for a substantial fraction of the
remaining confusion.

The present work develops that observation in a different direction. Rather
than adding further handcrafted geometric descriptors, we ask whether the
training objective itself can make use of empirically observed class
relationships. The underlying question is whether a classifier should devote
the same separation pressure to every possible rival class, or whether greater
pressure should be placed on class pairs that a preliminary model has already
demonstrated to be difficult.

Standard multiclass cross-entropy does not explicitly encode such pair-specific
structure. For an example with ground-truth class \(y\), the objective
encourages the model to favor \(y\) over competing classes, but it does not
distinguish between a persistent rival and a class that is already separated
with little difficulty. For example, a recurring I/L confusion and a nearly
nonexistent I/Z confusion participate in the same general multiclass objective
even though their practical difficulty is very different.

Margin-based representation learning provides a natural mechanism for imposing
additional discrimination. Contrastive and metric-learning methods explicitly
encourage separation between difficult examples or classes
\cite{hadsell2006dimensionality,schroff2015facenet}, while classification
methods such as Center Loss, CosFace, and ArcFace modify feature geometry to
increase discriminability
\cite{wen2016center,wang2018cosface,deng2019arcface}. Nevertheless, a single
uniform margin does not directly encode the empirical fact that some class
relationships are considerably more difficult than others.

Adaptive-margin methods have addressed related limitations in other settings.
AdaptiveFace learns class-dependent margins in response to unequal class
distributions \cite{liu2019adaptiveface}. MV-Softmax emphasizes misclassified
negative classes during optimization \cite{wang2020mvsoftmax}, while
CurricularFace changes the relative importance of easy and hard samples over
the course of training \cite{huang2020curricularface}. More recent work has
also incorporated confusion information directly into margin-based objectives,
including confusion-aware calibrated margins for vision-language model
adaptation \cite{wang2025cacm} and confusion-aware cosine constraints for
few-shot SAR recognition \cite{zhao2026confas}.

Accordingly, the contribution of the present study is not the broad claim that
adaptive or confusion-aware margins are themselves new. Instead, CAML studies
a specific formulation and experimental protocol for handwritten-letter
recognition: a \emph{sparse pair-specific confusion graph} is estimated from
a dedicated discovery partition, fixed before final-model training, and then
used to impose weighted true-class-versus-rival cosine constraints. The
proposed model is evaluated against a strictly matched baseline while preserving
the official test partition for final evaluation.

This separation is important. If difficult pairs are identified by inspecting
the final test confusion matrix and the model is subsequently modified in
response, the resulting test performance no longer represents a clean
assessment of generalization. CAML therefore separates four roles:
pilot training, confusion discovery, checkpoint selection, and final test
evaluation.

A short pilot convolutional neural network is first trained on a core subset
of the official EMNIST Letters training partition. A separate discovery
partition, held out from pilot training, is then used to estimate symmetric
pairwise confusion rates. The highest-ranked relationships form a sparse
weighted graph. After this graph has been fixed, the core and discovery
partitions are combined for final training of both the baseline and CAML
models. A separate validation partition is used for checkpoint selection, and
the official test partition is accessed only after model selection.

During CAML training, each selected graph edge introduces an additional
cosine-separation requirement between the true class and its discovered rival.
Let \(z\) denote the learned embedding of an input and let \(w_y\) and \(w_j\)
denote classifier weight vectors for the true class \(y\) and a selected rival
\(j\). CAML encourages

\[
\cos(z,w_y)-\cos(z,w_j)
\geq
m_{yj},
\]

where \(m_{yj}\) depends on the empirical strength of the discovered confusion
relationship. Pairs absent from the graph receive no additional CAML
constraint.

The method is intentionally evaluated as a controlled intervention. The
baseline and CAML networks use the same architecture, final training examples,
minibatch order, deterministic augmentation schedule, optimizer,
regularization, validation criterion, and early-stopping procedure. They also
contain the same number of trainable parameters. Their intended difference is
the presence or absence of the CAML auxiliary loss.

The central research question is therefore:

\begin{quote}
\emph{Can a strong augmented handwritten-letter classifier reduce persistent
errors between independently discovered rival classes by imposing additional
pair-specific representation-space margins on those relationships?}
\end{quote}

The executed experiment discovers eight principal confusion pairs,

\[
\{
\mathrm{I/L},
\mathrm{G/Q},
\mathrm{U/V},
\mathrm{R/V},
\mathrm{D/O},
\mathrm{C/E},
\mathrm{I/J},
\mathrm{H/N}
\}.
\]

Notably, the strongest discovered relationships are consistent with the
structured confusion observed in our earlier curvature--orientation study
\cite{nouri2025ijca,NouriCurv2025}, but the pairs used by CAML are not
manually transferred from that work. They are independently selected from the
dedicated discovery partition in the present experiment.

The resulting evidence is interpreted at two levels. At the global level,
clean-test accuracy changes only from \(95.6346\%\) to \(95.6731\%\), and the
paired difference is not statistically significant. Across ten corrupted
conditions, the average change is likewise small, and no individual
robustness comparison remains significant after correction for multiple
testing. At the targeted level, however, the total symmetric clean-test error
count over the eight selected confusion pairs decreases from 614 to 605.
The largest pair, I/L, improves from 344 to 334 errors, whereas the
second-largest pair, G/Q, worsens from 137 to 145.

These results distinguish three questions that are sometimes conflated:
whether an auxiliary objective is successfully optimized, whether it changes
the class relationships it targets, and whether those changes yield a
statistically reliable improvement in overall recognition. The first is clearly supported by the optimization behavior, the second is
supported only partially by the heterogeneous pair-level changes, and the
third is not established by the present experiment.

The main contributions of this study are therefore:

\begin{itemize}[leftmargin=1.5em]

    \item We extend our earlier curvature--orientation handwriting-recognition
    work \cite{nouri2025ijca,NouriCurv2025} from explicit geometric feature
    representation to confusion-aware representation learning.

    \item We formulate \textbf{Confusion-Adaptive Margin Learning (CAML)}
    using a sparse, weighted, pair-specific confusion graph discovered from
    data before final-model training.

    \item We use a \textbf{leakage-controlled experimental protocol} that
    separates pilot training, pair discovery, final validation, and official
    test evaluation.

    \item We isolate the effect of the proposed loss through a
    \textbf{strictly matched baseline comparison}: architecture, training
    examples, augmentation sequence, optimizer, regularization, and model
    selection are held fixed.

    \item We evaluate both \textbf{aggregate recognition performance} and
    \textbf{targeted pair behavior}, allowing the effect of the margin
    objective to be distinguished from changes in overall top-1 accuracy.

    \item We use paired McNemar testing and Holm correction for the robustness
    comparisons, and we interpret the resulting small differences
    conservatively rather than treating directional gains as established
    superiority.

\end{itemize}

The broader implication is that a confusion matrix can identify
\emph{where} a classifier remains uncertain without necessarily revealing
\emph{how} the representation should be modified to resolve that uncertainty.
A useful confusion-aware method must therefore be evaluated not only by whether
its targeted constraints become easier to satisfy, but also by whether the
resulting changes improve predictions without introducing compensating errors
elsewhere.

\section{Related Work}
\label{sec:related_work}

\subsection{Handwritten Character Recognition}

Handwritten-character recognition has progressed from manually engineered
geometric and statistical descriptors to neural models that learn
representations directly from images. Convolutional architectures became a
standard approach to handwritten recognition because they preserve local
spatial organization while progressively combining low-level stroke
information into higher-level representations \cite{lecun1998gradient}.

The EMNIST benchmark extended the MNIST setting to handwritten letters and
larger alphanumeric classification problems \cite{cohen2017emnist}. EMNIST
Letters contains 26 balanced alphabet classes and is particularly suitable for
studying fine-grained confusion because several letters possess closely
related handwritten forms.

Our earlier work approached this problem from the representation side,
investigating planar curvature and gradient orientation as explicit geometric
features for handwritten recognition
\cite{nouri2025ijca,NouriCurv2025}. The present study instead focuses on
residual class relationships after a strong learned representation has already
been obtained. This shift motivates an objective that reacts to empirical
confusion structure rather than enriching every input with additional
handcrafted features.

\subsection{Metric and Margin-Based Representation Learning}

Metric learning seeks representations in which semantically related examples
are close while unrelated examples are separated. Contrastive loss imposes
pairwise distance constraints \cite{hadsell2006dimensionality}, while FaceNet
uses triplet-based learning to separate an anchor from a difficult negative
relative to a positive example \cite{schroff2015facenet}.

Related classification objectives modify the geometry of the representation
space while retaining a multiclass classifier. Center Loss encourages
within-class compactness by penalizing distance from class-dependent feature
centers \cite{wen2016center}. CosFace introduces an additive cosine margin
between the target class and competing classes \cite{wang2018cosface}, and
ArcFace imposes an additive angular margin
\cite{deng2019arcface}. These methods demonstrate that classification
performance can be influenced by explicitly controlling feature--class
geometry.

A limitation of globally fixed margins is that they apply essentially the same
separation principle across classes even when the empirical difficulty of
those classes differs. AdaptiveFace addresses one aspect of this problem by
learning class-dependent margins, motivated primarily by imbalance in the
amount of data available for different identities
\cite{liu2019adaptiveface}. CAML differs in that its adaptation is not based
on class frequency and is not represented by one margin per class. Instead,
the additional constraints are attached to particular class-pair edges
identified through observed confusion.

Hard-example methods provide another related direction. MV-Softmax
specifically increases emphasis on misclassified negative classes rather than
treating all non-target classes identically \cite{wang2020mvsoftmax}.
CurricularFace further adjusts the relative emphasis on easy and hard examples
according to the stage of training \cite{huang2020curricularface}. These
approaches adapt supervision according to sample or optimization difficulty,
whereas CAML first estimates a persistent class-level relationship on separate
data and then keeps that sparse structural prior fixed during final training.

\subsection{Confusion-Aware Learning}

A confusion matrix contains information that is lost in aggregate accuracy:
it identifies which true and predicted classes repeatedly interact. This makes
confusion statistics a natural source of supervision for targeted learning.
However, the way confusion information is obtained is methodologically
important. If final-test errors determine which class relationships receive
special treatment, subsequent evaluation on the same test set is no longer
fully independent.

Recent work has incorporated confusion information into adaptive learning
objectives in several settings. Wang et al.\ introduce a confusion-aware
calibrated margin within a vision-language model adaptation framework designed
to address imbalanced pseudolabels \cite{wang2025cacm}. Their method uses
concept-confusion information to calibrate learning under a pseudolabel
adaptation setting. This establishes that confusion-derived margin adjustment
is useful beyond conventional fixed-margin classification, but its objective,
data regime, and motivation differ from the supervised handwritten-character
setting considered here.

More closely related, Zhao et al.\ introduce a confusion-aware cosine loss
within ConFAS-Net for few-shot synthetic-aperture-radar target recognition
\cite{zhao2026confas}. Their framework dynamically derives confusion
information during a multi-component few-shot recognition procedure and
combines the confusion-aware loss with attention and decision-adjustment
modules. CAML instead isolates the confusion-aware margin mechanism in a
single-model supervised experiment. It constructs a fixed sparse graph of
specific rival relationships from a disjoint discovery partition before final
training and introduces no additional attention module, decision-adjustment
stage, or inference-time mechanism.

These distinctions define the intended contribution of CAML more precisely.
The method is not presented as the first use of adaptive margins or confusion
statistics. Rather, it studies whether a \emph{pre-discovered sparse
pair-specific confusion graph} can be translated into weighted cosine
constraints under a controlled, leakage-aware protocol, and whether satisfying
those targeted constraints produces measurable improvement in the associated
class relationships and in overall recognition.

A second distinction is that CAML does not create pair-specific specialist
classifiers. The confusion graph affects only optimization. At inference time,
a single network still produces one 26-class logit vector, and the graph is
neither consulted nor used for routing. This allows any observed difference to
be attributed to changes induced during training rather than to additional
test-time machinery.

\section{Confusion-Adaptive Margin Learning}
\label{sec:method}

\subsection{Problem Formulation}

Let

\[
\mathcal{D}
=
\left\{
(x_n,y_n)
\right\}_{n=1}^{N},
\qquad
y_n \in \{0,\ldots,C-1\},
\]

denote a labeled handwritten-character dataset with \(C=26\) classes.
A convolutional network maps an input image \(x\) to a \(d\)-dimensional
embedding

\[
z=f_{\theta}(x)\in\R^d.
\]

The embedding is passed to a linear classifier with class-weight vectors
\(w_j\) and biases \(b_j\). The logit for class \(j\) is

\begin{equation}
h_j(x)
=
w_j^\top z+b_j,
\label{eq:logit}
\end{equation}

and the corresponding class probability is

\begin{equation}
p(j\mid x)
=
\frac{\exp(h_j(x))}
{\displaystyle\sum_{k=0}^{C-1}\exp(h_k(x))}.
\label{eq:softmax}
\end{equation}

The conventional classification objective is cross-entropy. In the
experiments reported here, label smoothing is used with smoothing coefficient
\(\epsilon_{\mathrm{LS}}\). For an example \(n\), the smoothed target
distribution is

\begin{equation}
q_{nk}
=
(1-\epsilon_{\mathrm{LS}})
\mathbb{I}[k=y_n]
+
\frac{\epsilon_{\mathrm{LS}}}{C},
\label{eq:smoothed_target}
\end{equation}

where \(\mathbb{I}[\cdot]\) is the indicator function. The minibatch
classification loss is therefore

\begin{equation}
\mathcal{L}_{\mathrm{CE}}
=
-\frac{1}{B}
\sum_{n=1}^{B}
\sum_{k=0}^{C-1}
q_{nk}
\log p(k\mid x_n),
\label{eq:ce}
\end{equation}

where \(B\) is the minibatch size. In the present experiments,

\[
\epsilon_{\mathrm{LS}}=0.05.
\]

CAML does not replace this multiclass classification objective. Instead, it
adds an auxiliary constraint for particular rival class pairs identified
before final-model training by an independently trained pilot network.

\subsection{Leakage-Controlled Data Partitioning}

The official EMNIST Letters training partition contains 124,800 examples, and
the official test partition contains 20,800 examples
\cite{cohen2017emnist}. To separate confusion discovery from final model
selection and test evaluation, the official training partition is divided into
three mutually disjoint subsets:

\begin{enumerate}[leftmargin=1.6em]

    \item \textbf{Core set:} 106,080 examples, corresponding to \(85\%\) of
    the official training partition, used to train the pilot model.

    \item \textbf{Discovery set:} 6,240 examples, corresponding to \(5\%\),
    held out from pilot training and used only to estimate the confusion
    structure that determines the CAML graph.

    \item \textbf{Validation set:} 12,480 examples, corresponding to \(10\%\),
    reserved for checkpoint selection of the two final models.

\end{enumerate}

After the confusion graph has been fixed, the core and discovery subsets are
combined to form the 112,320-example final training set. The same final
training examples are used for both the cross-entropy baseline and CAML.

Thus, the discovery subset is \emph{held out from pilot training}, but it is
not permanently excluded from optimization: after its sole role in graph
construction has been completed, it becomes part of the common final training
set. The validation set is never used for pair discovery, and the official
test partition is not used for graph construction or checkpoint selection.

This design separates four distinct operations:

\begin{enumerate}[leftmargin=1.6em]
    \item learning a preliminary classifier;
    \item estimating persistent class-pair confusion;
    \item selecting final-model checkpoints; and
    \item evaluating the selected models on the official test partition.
\end{enumerate}

\begin{figure}[H]
    \centering
    \includegraphics[width=0.94\textwidth]{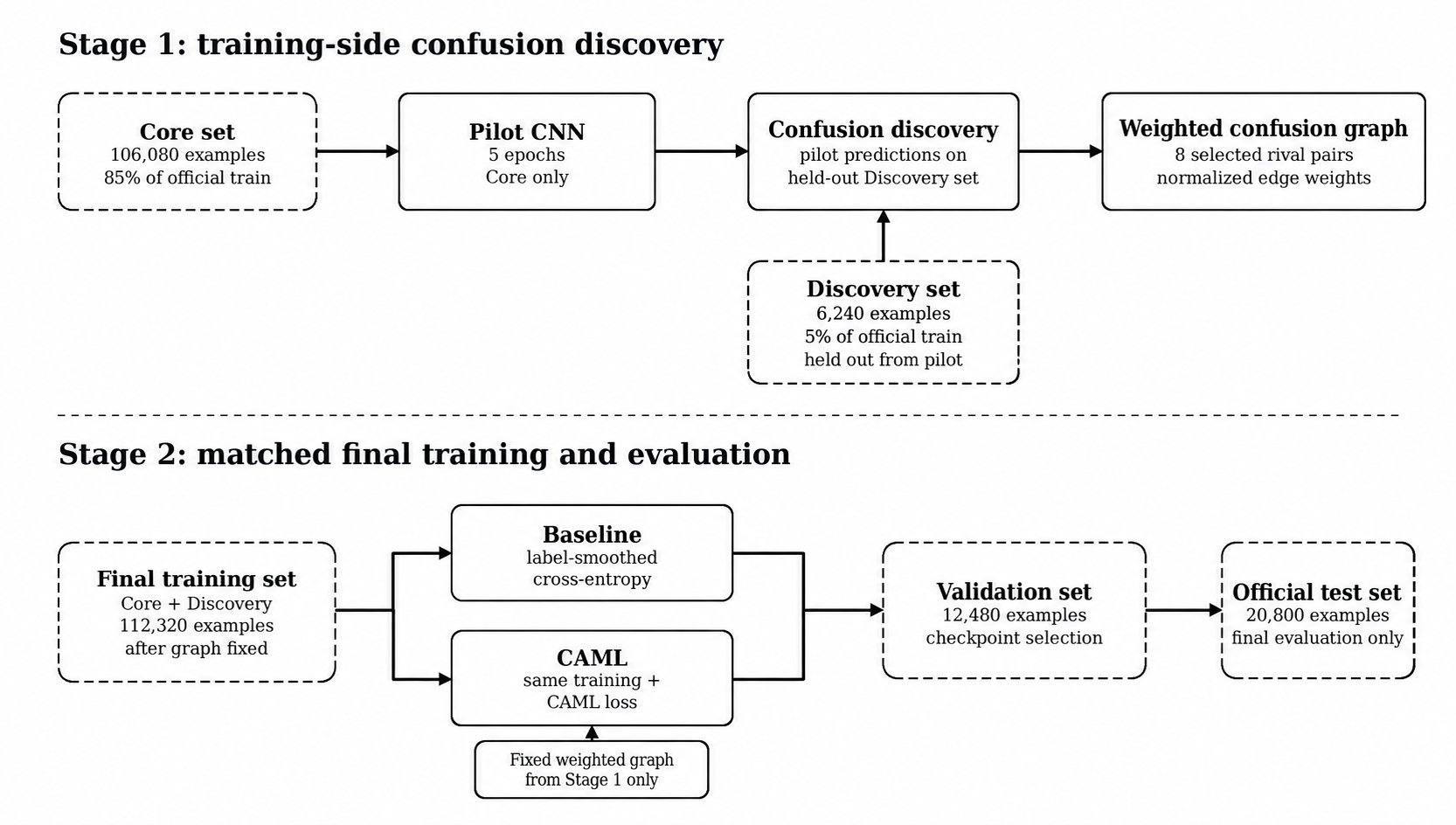}
    \caption{
    Leakage-controlled CAML experimental protocol.
    In Stage 1, the pilot network is trained only on the core subset and evaluated
    on the separate discovery subset to construct the weighted confusion graph.
    After the graph is fixed, the core and discovery subsets are recombined into
    the common final training set. In Stage 2, the baseline and CAML models are
    trained under matched conditions; the confusion graph affects only the CAML
    objective. The validation subset is used for checkpoint selection, and the
    official EMNIST Letters test partition is reserved for final evaluation.
    }
    \label{fig:caml_protocol}
\end{figure}

Figure~\ref{fig:caml_protocol} summarizes the complete leakage-controlled
experimental protocol and the distinct roles assigned to the core, discovery,
validation, and official test partitions.


\subsection{Pilot Network and Confusion Discovery}

A pilot convolutional network is trained for five epochs using only the core
subset. The pilot architecture is identical to that used by the final models.
Its purpose is not to maximize final recognition accuracy, but to obtain a
sufficiently strong classifier from which persistent confusion relationships
can be estimated on previously unseen discovery examples.

Let

\[
M\in\mathbb{N}^{C\times C}
\]

denote the confusion matrix obtained by evaluating the pilot network on the
discovery partition. Entry \(M_{ij}\) is the number of discovery examples
whose true class is \(i\) and whose predicted class is \(j\).

For every unordered pair \((i,j)\), \(i<j\), we compute the symmetric confusion
count

\begin{equation}
E_{ij}
=
M_{ij}+M_{ji},
\label{eq:sym_count}
\end{equation}

and the symmetric confusion rate

\begin{equation}
r_{ij}
=
\frac{1}{2}
\left(
\frac{M_{ij}}{N_i}
+
\frac{M_{ji}}{N_j}
\right),
\label{eq:sym_rate}
\end{equation}

where \(N_i\) and \(N_j\) denote the numbers of discovery examples belonging
to classes \(i\) and \(j\), respectively.

Pairs with fewer than five symmetric errors are excluded. The remaining pairs
are ranked primarily by \(r_{ij}\), with the total symmetric error count
\(E_{ij}\) used as a secondary criterion. The eight highest-ranked eligible
pairs define the sparse confusion graph used during final CAML training.

For the executed experiment, the resulting graph contains

\[
\{
\mathrm{I/L},
\mathrm{G/Q},
\mathrm{U/V},
\mathrm{R/V},
\mathrm{D/O},
\mathrm{C/E},
\mathrm{I/J},
\mathrm{H/N}
\}.
\]

Each selected edge receives a normalized confusion weight

\begin{equation}
\alpha_{ij}
=
\frac{r_{ij}}
{\displaystyle\max_{(u,v)\in\mathcal{E}} r_{uv}},
\label{eq:edge_weight}
\end{equation}

where \(\mathcal{E}\) denotes the set of selected graph edges. The strongest
discovered pair therefore receives weight one, while weaker selected
relationships receive proportionally smaller weights.

\subsection{Confusion-Adaptive Cosine Margin}
\label{subsec:caml_margin}

After discovery is complete, the selected relationships are stored in a
symmetric matrix

\[
A\in\R^{C\times C},
\]

defined by

\begin{equation}
A_{ij}
=
\begin{cases}
\alpha_{ij}, & (i,j)\in\mathcal{E},\\
0,           & \text{otherwise}.
\end{cases}
\label{eq:adjacency}
\end{equation}

Only pairs represented by nonzero entries of \(A\) contribute to the CAML
auxiliary objective.

For example \(x_n\), let

\[
z_n=f_{\theta}(x_n)
\]

be its learned embedding, and let \(w_j\) be the weight vector of classifier
class \(j\). The standard classifier continues to use the full affine logit
\(w_j^\top z_n+b_j\) from Eq.~\eqref{eq:logit}. CAML, however, operates on
normalized embeddings and classifier weight directions:

\begin{equation}
\hat{z}_n
=
\frac{z_n}{\|z_n\|_2},
\qquad
\hat{w}_j
=
\frac{w_j}{\|w_j\|_2}.
\label{eq:normalization}
\end{equation}

The cosine similarity between example \(n\) and class direction \(j\) is

\begin{equation}
s_{nj}
=
\hat{z}_n^\top\hat{w}_j.
\label{eq:cos_sim}
\end{equation}

For an example whose true class is \(y_n\), CAML requires the similarity to
the true-class direction to exceed the similarity to a selected rival \(j\)
by a positive margin:

\begin{equation}
s_{n y_n}-s_{nj}
\geq
m_{y_nj}.
\label{eq:margin_requirement}
\end{equation}

The required pair-specific margin is

\begin{equation}
m_{ij}
=
m_0+m_1A_{ij},
\label{eq:adaptive_margin}
\end{equation}

where \(m_0\) is the base margin and \(m_1\) controls the additional margin
assigned according to the strength of the discovered confusion relationship.

The executed experiment uses

\[
m_0=0.10,
\qquad
m_1=0.10.
\]

Consequently, selected edges receive margins between slightly above \(0.10\)
and a maximum of

\[
m_{\max}
=
0.10+0.10(1)
=
0.20.
\]

This construction differs from a globally uniform large-margin objective.
Additional separation is requested only for class relationships selected by
the discovery procedure, and the magnitude of the requested separation
depends on the observed confusion strength.

\subsection{Pair-Weighted Margin Loss}

For a training example from class \(y_n\) and a selected rival \(j\), a hinge
penalty is incurred whenever Eq.~\eqref{eq:margin_requirement} is violated:

\begin{equation}
\ell_{nj}
=
A_{y_nj}
\left[
m_{y_nj}
-
\left(
s_{ny_n}-s_{nj}
\right)
\right]_{+},
\label{eq:hinge_pair}
\end{equation}

where

\[
[a]_{+}=\max(0,a).
\]

The factor \(A_{y_nj}\) serves both as an activation mask and as an edge
weight. Rival classes outside the discovered graph contribute zero penalty,
while stronger selected relationships receive greater weight.

For minibatch \(\mathcal{B}\), the CAML auxiliary loss is

\begin{equation}
\mathcal{L}_{\mathrm{CAML}}
=
\frac{
\displaystyle
\sum_{n\in\mathcal{B}}
\sum_{j=0}^{C-1}
A_{y_nj}
\left[
m_{y_nj}
-
\left(
s_{ny_n}-s_{nj}
\right)
\right]_{+}
}{
\displaystyle
\sum_{n\in\mathcal{B}}
\sum_{j=0}^{C-1}
A_{y_nj}
+
\varepsilon
},
\label{eq:caml_loss}
\end{equation}

where \(\varepsilon\) is a small numerical constant that prevents division by
zero in minibatches containing no active graph relationships.

The final CAML training objective is

\begin{equation}
\boxed{
\mathcal{L}
=
\mathcal{L}_{\mathrm{CE}}
+
\lambda
\mathcal{L}_{\mathrm{CAML}}
}
\label{eq:total_loss}
\end{equation}

with

\[
\lambda=0.20.
\]

The ordinary label-smoothed cross-entropy objective therefore remains
responsible for learning the complete 26-class task. CAML acts only as an
auxiliary representation-space constraint on class relationships identified
by the discovery procedure.

\subsection{Inference}

CAML modifies training but does not alter the prediction rule. Given a test
image \(x\), the network produces 26 logits according to
Eq.~\eqref{eq:logit}, and the predicted class is

\begin{equation}
\hat{y}
=
\arg\max_j h_j(x).
\label{eq:prediction}
\end{equation}

The confusion graph is not consulted during inference. No pair-specific
classifier, confidence threshold, post-hoc correction, or routing mechanism is
used. CAML therefore has the same parameter count and prediction-time
architecture as the matched cross-entropy baseline.

\section{Network Architecture}
\label{sec:architecture}

To isolate the effect of the training objective, the pilot, baseline, and CAML
models use the same convolutional architecture. The network processes a
single-channel \(28\times28\) image using stacked convolutional layers,
batch normalization \cite{ioffe2015batch}, GELU activations, max pooling, and
dropout.

Each convolution uses a \(3\times3\) kernel. The first three convolutional
layers operate with 32 channels, followed by two 64-channel layers and two
128-channel layers. Spatial resolution is reduced through two
\(2\times2\) max-pooling operations.

After the convolutional trunk, the \(128\times7\times7\) feature map is
flattened and projected to a 256-dimensional embedding. This embedding is the
representation used by the CAML cosine-separation objective. A final dropout
layer precedes the 26-class affine classifier.

Table~\ref{tab:architecture} summarizes the complete network.

\begin{table}[htbp]
\centering
\caption{CNN architecture used by the pilot, cross-entropy baseline, and CAML model.}
\label{tab:architecture}
\setlength{\tabcolsep}{5pt}
\renewcommand{\arraystretch}{1.25}
\begin{tabular}{lll}
\toprule
\textbf{Stage} & \textbf{Output size} & \textbf{Operation} \\
\midrule

Input
& \(1\times28\times28\)
& Grayscale image \\

Stem 1
& \(32\times28\times28\)
& \(3\times3\) Conv, BN, GELU \\

Stem 2
& \(32\times28\times28\)
& \(3\times3\) Conv, BN, GELU \\

Adapter
& \(32\times28\times28\)
& \(3\times3\) Conv, BN, GELU \\

Pool 1
& \(32\times14\times14\)
& \(2\times2\) max pooling \\

Conv block 1
& \(64\times14\times14\)
& \(3\times3\) Conv, BN, GELU \\

Conv block 2
& \(64\times14\times14\)
& \(3\times3\) Conv, BN, GELU \\

Regularization
& \(64\times14\times14\)
& Spatial dropout \(p=0.10\) \\

Pool 2
& \(64\times7\times7\)
& \(2\times2\) max pooling \\

Conv block 3
& \(128\times7\times7\)
& \(3\times3\) Conv, BN, GELU \\

Conv block 4
& \(128\times7\times7\)
& \(3\times3\) Conv, BN, GELU \\

Regularization
& \(128\times7\times7\)
& Spatial dropout \(p=0.15\) \\

Embedding
& \(256\)
& Flatten, FC, GELU \\

Classifier input
& \(256\)
& Dropout \(p=0.35\) \\

Output
& \(26\)
& Affine linear classifier \\

\bottomrule
\end{tabular}
\end{table}

Each network contains

\[
1{,}908{,}730
\]

trainable parameters, including the classifier biases. CAML introduces no
additional trainable module relative to the baseline. Its only modification is
the auxiliary loss applied to the existing 256-dimensional embedding and
classifier weight vectors during training.

\FloatBarrier
\section{Experimental Protocol}
\label{sec:experimental_protocol}

\subsection{Dataset}

Experiments use the official EMNIST Letters dataset
\cite{cohen2017emnist}. The task contains 26 balanced handwritten-letter
classes. The original labels \(1,\ldots,26\) are shifted to

\[
0,\ldots,25,
\]

corresponding to A--Z.

The official partition contains 124,800 training examples and 20,800 test
examples. The official test partition is preserved throughout the experiment
rather than merged with the training data and resplit.

Table~\ref{tab:data_split} summarizes the complete data protocol.

\begin{table}[htbp]
\centering
\caption{Partitioning used in the CAML experiment. The discovery subset is
held out from pilot training and is recombined with the core subset only after
the confusion graph has been fixed.}
\label{tab:data_split}
\setlength{\tabcolsep}{6pt}
\renewcommand{\arraystretch}{1.25}
\begin{tabular}{lrrl}
\toprule
\textbf{Partition} &
\textbf{Examples} &
\textbf{Fraction of official train} &
\textbf{Purpose} \\
\midrule

Core
& 106,080
& 85\%
& Pilot training \\

Discovery
& 6,240
& 5\%
& Confusion-graph construction \\

Validation
& 12,480
& 10\%
& Final checkpoint selection \\

\midrule

Final train
& 112,320
& 90\%
& Core + discovery \\

Official test
& 20,800
& --
& Final evaluation only \\

\bottomrule
\end{tabular}
\end{table}

The core, discovery, and validation subsets are mutually disjoint. Confusion
discovery does not use the final validation set, and neither graph construction
nor checkpoint selection uses the official test partition.

\subsection{Data Augmentation}
\label{subsec:augmentation}

A strong augmented baseline is used so that CAML is evaluated against ordinary
robustness-oriented training rather than against an intentionally weak
reference model. The baseline and CAML models receive the same deterministic
augmentation schedule.

For each training example, one augmentation category is sampled according to

\begin{equation}
P(\text{augmentation})
=
(0.30,\,0.12,\,0.12,\,0.12,\,0.12,\,0.11,\,0.11),
\label{eq:aug_probs}
\end{equation}

corresponding respectively to

\[
\{
\text{clean},
\text{rotation},
\text{translation},
\text{noise},
\text{blur},
\text{thickening},
\text{thinning}
\}.
\]

Only one augmentation family is applied to an example during a given training
presentation. The transformations are defined as follows:

\begin{itemize}[leftmargin=1.5em]

    \item \textbf{Rotation:}
    an angle sampled uniformly from
    \([-10^\circ,10^\circ]\).

    \item \textbf{Translation:}
    horizontal and vertical integer shifts independently sampled from
    \[
    \{-2,-1,0,1,2\}.
    \]

    \item \textbf{Gaussian noise:}
    additive zero-mean noise with standard deviation sampled uniformly from
    \[
    [0.02,0.12],
    \]
    after image intensities have been normalized to \([0,1]\).

    \item \textbf{Gaussian blur:}
    standard deviation selected from
    \[
    \sigma=0.6
    \quad\text{or}\quad
    \sigma=1.0.
    \]

    \item \textbf{Stroke thickening:}
    a \(3\times3\) max operation that expands foreground strokes.

    \item \textbf{Stroke thinning:}
    a corresponding \(3\times3\) erosion operation.

    \item \textbf{Clean:}
    the normalized image is retained without an additional perturbation.

\end{itemize}

The augmentation sequence is deterministically seeded by epoch and minibatch.
Consequently, the baseline and CAML models receive the same minibatch order and
the same transformed examples at corresponding optimization steps. This
controls augmentation randomness as a source of systematic difference between
the models.

\subsection{Optimization}

Both final networks are trained from scratch using AdamW with initial learning
rate

\[
\eta=10^{-3}
\]

and weight decay

\[
10^{-4}.
\]

Training uses minibatches of 256 examples and the label-smoothed cross-entropy
objective defined in Eq.~\eqref{eq:ce}, with

\[
\epsilon_{\mathrm{LS}}=0.05.
\]

The maximum training duration is 30 epochs. A
\texttt{ReduceLROnPlateau} scheduler monitors the validation criterion and
reduces the learning rate by a factor of \(0.5\) after two plateau epochs,
with minimum learning rate \(10^{-6}\).

Early stopping uses patience six. The best checkpoint for each model is
restored according to the mean accuracy across the fixed validation suite
described below rather than according to clean validation accuracy alone.

For CAML,

\[
\lambda=0.20,
\qquad
m_0=0.10,
\qquad
m_1=0.10.
\]

The baseline uses the identical architecture and optimization protocol but
contains no CAML auxiliary term; equivalently,

\[
\lambda=0.
\]

\subsection{Validation Criterion}

Checkpoint selection is based on equally weighted mean accuracy across seven
fixed validation conditions:

\begin{enumerate}[leftmargin=1.6em]
    \item clean images;
    \item random rotation up to \(10^\circ\);
    \item translation up to two pixels;
    \item Gaussian noise with \(\sigma=0.10\);
    \item Gaussian blur with \(\sigma=1.0\);
    \item one-step stroke thickening; and
    \item one-step stroke thinning.
\end{enumerate}

If \(a_k\) denotes validation accuracy under condition \(k\), the checkpoint
criterion is

\begin{equation}
S_{\mathrm{val}}
=
\frac{1}{7}
\sum_{k=1}^{7}
a_k.
\label{eq:validation_score}
\end{equation}

This criterion prevents checkpoint selection from being based exclusively on
the clean validation distribution.

\subsection{Test Protocol and Statistical Evaluation}
\label{subsec:test_protocol}

After checkpoint selection has been completed, the baseline and CAML models
are evaluated on the 20,800-example official EMNIST Letters test partition.

In addition to clean images, the final evaluation includes ten predefined
perturbation conditions:

\[
\begin{aligned}
&\text{rotation: } 5^\circ,\ 10^\circ,\\
&\text{translation: } 1,\ 2\text{ pixels},\\
&\text{Gaussian noise: } \sigma=0.05,\ 0.10,\\
&\text{Gaussian blur: } \sigma=0.6,\ 1.0,\\
&\text{stroke thickening: } 1\text{ step},\\
&\text{stroke thinning: } 1\text{ step}.
\end{aligned}
\]

The use of controlled image corruptions provides a complementary robustness
view rather than treating clean accuracy as the sole evaluation criterion
\cite{hendrycks2019benchmarking}. Top-1 accuracy is the principal metric, and
macro-averaged F1 is additionally reported for the clean test partition.

Because the baseline and CAML models are evaluated on exactly the same
examples, their predictions are paired. For each condition, we therefore use
the exact McNemar test \cite{mcnemar1947note}. Let

\[
n_{01}
=
\#\{
\text{baseline wrong, CAML correct}
\}
\]

and

\[
n_{10}
=
\#\{
\text{baseline correct, CAML wrong}
\}.
\]

The test evaluates whether the two discordant transition counts are consistent
with equal paired error probabilities.

For the ten non-clean perturbation conditions, the resulting \(p\)-values are
adjusted using the sequential Holm procedure
\cite{holm1979simple} to control the family-wise error rate associated with
multiple comparisons.

The clean-test McNemar comparison is reported separately because it is the
principal uncorrupted evaluation rather than one member of the ten-condition
robustness family. Pair-specific confusion counts are also reported
descriptively because CAML explicitly targets those relationships; however,
no separate inferential significance claim is made for the individual
pair-level reductions.

\section{Confusion-Graph Discovery}
\label{sec:discovery_results}

The five-epoch pilot CNN achieves \(95.3365\%\) accuracy on the held-out
discovery partition. Its residual errors are highly nonuniform across class
pairs.

Table~\ref{tab:discovered_pairs} reports the eight relationships selected by
the confusion-discovery procedure. The directional columns give the raw number
of pilot errors in each direction. The symmetric error count is the sum of the
two directional counts, while the symmetric confusion rate is defined in
Eq.~\eqref{eq:sym_rate}.

\begin{table}[htbp]
\centering
\caption{Confusion pairs discovered from the discovery partition. Neither the
final validation set nor the official test set is used to construct this table.}
\label{tab:discovered_pairs}
\setlength{\tabcolsep}{4pt}
\renewcommand{\arraystretch}{1.25}
\begin{tabular}{lrrrrr}
\toprule
\textbf{Pair} &
\textbf{First$\rightarrow$Second} &
\textbf{Second$\rightarrow$First} &
\textbf{Sym. errors} &
\textbf{Sym. rate} &
\textbf{Weight} \\
\midrule

I/L & 38 & 66 & 104 & 0.2167 & 1.0000 \\
G/Q & 24 & 14 & 38  & 0.0792 & 0.3654 \\
U/V & 3  & 6  & 9   & 0.0188 & 0.0865 \\
R/V & 8  & 1  & 9   & 0.0188 & 0.0865 \\
D/O & 7  & 1  & 8   & 0.0167 & 0.0769 \\
C/E & 4  & 3  & 7   & 0.0146 & 0.0673 \\
I/J & 3  & 4  & 7   & 0.0146 & 0.0673 \\
H/N & 4  & 2  & 6   & 0.0125 & 0.0577 \\

\bottomrule
\end{tabular}
\end{table}

\begin{figure}[htbp]
\centering

\resizebox{0.94\textwidth}{!}{%
\begin{tikzpicture}[
    letter/.style={
        circle,
        draw,
        minimum size=9mm,
        inner sep=0pt,
        font=\bfseries\small
    },
    weight/.style={
        fill=white,
        inner sep=1.5pt,
        font=\scriptsize
    }
]


\node[letter] (I) at (0,0) {I};
\node[letter] (L) at (-2.0,1.0) {L};
\node[letter] (J) at (-2.0,-1.0) {J};

\draw[line width=3.6pt]
(I) -- node[weight,above left]{1.0000} (L);

\draw[line width=1.1pt]
(I) -- node[weight,below left]{0.0673} (J);


\node[letter] (G) at (2.3,1.0) {G};
\node[letter] (Q) at (4.4,1.0) {Q};

\draw[line width=2.2pt]
(G) -- node[weight,above]{0.3654} (Q);


\node[letter] (U) at (2.3,-1.0) {U};
\node[letter] (V) at (4.4,-1.0) {V};
\node[letter] (R) at (6.5,-1.0) {R};

\draw[line width=1.3pt]
(U) -- node[weight,above]{0.0865} (V);

\draw[line width=1.3pt]
(V) -- node[weight,above]{0.0865} (R);


\node[letter] (D) at (-1.0,-3.2) {D};
\node[letter] (O) at (1.1,-3.2) {O};

\draw[line width=1.2pt]
(D) -- node[weight,above]{0.0769} (O);


\node[letter] (C) at (2.8,-3.2) {C};
\node[letter] (E) at (4.9,-3.2) {E};

\draw[line width=1.1pt]
(C) -- node[weight,above]{0.0673} (E);


\node[letter] (H) at (6.6,-3.2) {H};
\node[letter] (N) at (8.7,-3.2) {N};

\draw[line width=1.0pt]
(H) -- node[weight,above]{0.0577} (N);

\end{tikzpicture}
}

\caption{
Sparse confusion graph discovered from the dedicated discovery partition.
Nodes denote letter classes and edges denote the eight selected rival
relationships. Edge labels are the normalized confusion weights
\(\alpha_{ij}\). The dominant I/L relationship receives weight \(1.0\),
followed by G/Q with weight \(0.3654\); the remaining selected relationships
receive substantially smaller weights. The graph is fixed before final CAML
training and is not used for inference.
}
\label{fig:confusion_graph}

\end{figure}
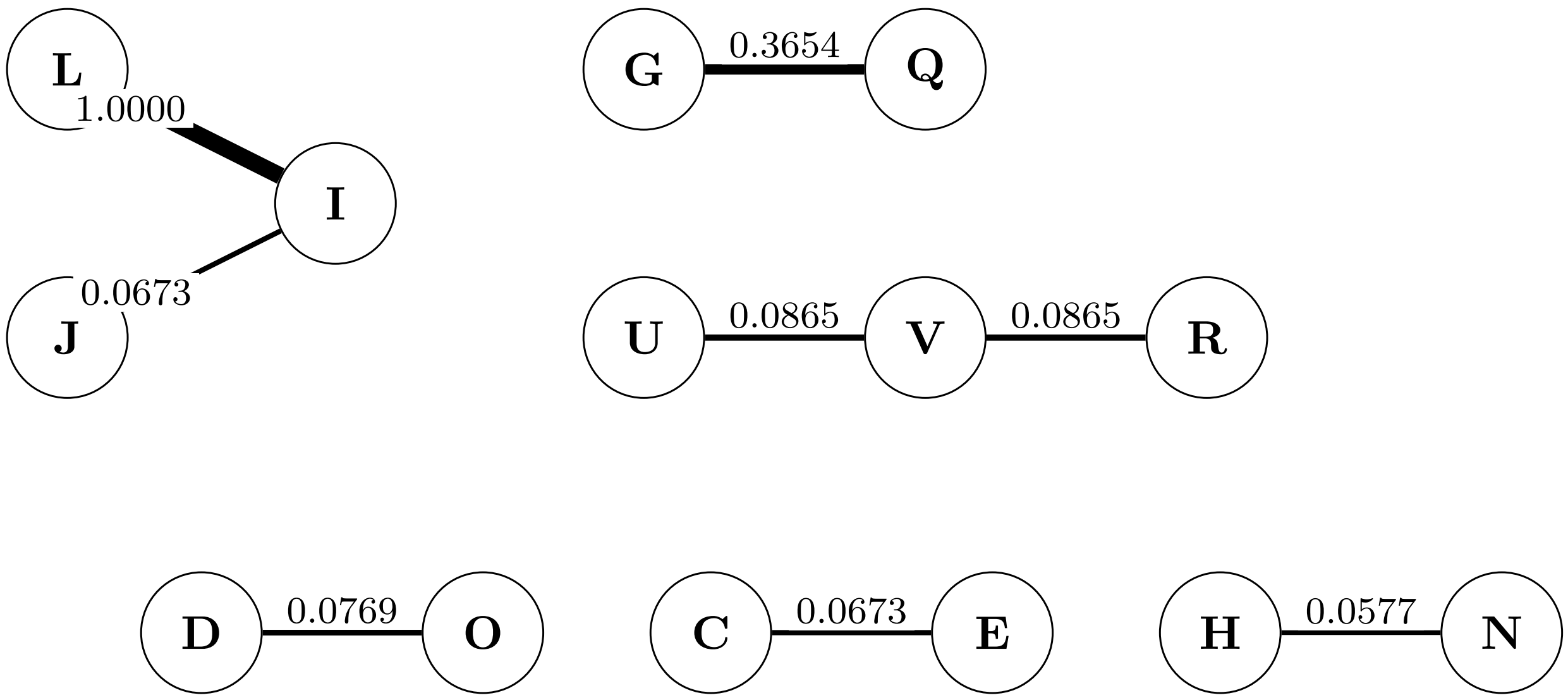

Figure~\ref{fig:confusion_graph} provides a graphical view of the same
discovered relationships. Edge labels show the normalized confusion weights,
while line thickness emphasizes the relative strength of the selected edges.

Two properties are immediately visible. First, the graph is highly
concentrated. I/L accounts for 104 symmetric discovery errors and therefore
receives the maximum normalized weight. G/Q is the second-largest relationship
with 38 errors and weight \(0.3654\). Every remaining selected edge contains
fewer than ten symmetric discovery errors.

Second, confusion can be directionally asymmetric even though CAML represents
the selected relationship by a symmetric graph edge. For I/L, the pilot
predicts I as L 38 times but L as I 66 times. Both directions contribute to
the pair-selection statistic, and once the pair is selected, examples from
either class are encouraged to separate from the other class during final
training.

The two strongest discovered relationships, I/L and G/Q, are also consistent
with prominent confusions observed in our earlier curvature--orientation study
\cite{nouri2025ijca,NouriCurv2025}. Importantly, they were not manually
transferred from that work: the pair list used by CAML is determined
independently from the present discovery partition.

The resulting graph should therefore be interpreted as a sparse empirical
description of the class relationships that remain difficult for the pilot
classifier. It is used only to define the auxiliary training objective and
does not alter the output label space or introduce any test-time routing.

\section{Results}
\label{sec:results}

\subsection{Validation Performance and Checkpoint Selection}

The two final models are selected independently using the robustness-averaged
validation criterion in Eq.~\eqref{eq:validation_score}. The cross-entropy
baseline reaches its highest mean validation score at epoch 24, with
\(95.5151\%\), whereas CAML reaches its highest mean validation score at epoch
27, with \(95.5117\%\).

The difference between the selected validation scores is therefore only
\(-0.0034\) percentage points in favor of the baseline. This near equality is
important because the CAML checkpoint is not selected on the basis of a clear
validation advantage.

Table~\ref{tab:best_validation} reports the condition-specific validation
accuracies at the selected checkpoints.

\begin{table}[htbp]
\centering
\caption{Validation accuracy at the independently selected best checkpoints.
Checkpoint selection uses the mean across all seven conditions.}
\label{tab:best_validation}
\setlength{\tabcolsep}{5pt}
\renewcommand{\arraystretch}{1.20}
\begin{tabular}{lrrr}
\toprule
\textbf{Condition} &
\textbf{Baseline (\%)} &
\textbf{CAML (\%)} &
\textbf{CAML--Base. (pp)} \\
\midrule

Clean                  & 95.8814 & 95.8814 &  0.0000 \\
Rotation \(10^\circ\)  & 95.4567 & 95.5369 & +0.0801 \\
Translation 2 px       & 95.3446 & 95.3365 & -0.0080 \\
Noise \(\sigma=0.10\)  & 95.6090 & 95.5689 & -0.0401 \\
Blur \(\sigma=1.0\)    & 95.7212 & 95.5610 & -0.1603 \\
Stroke thickening      & 95.6170 & 95.6811 & +0.0641 \\
Stroke thinning        & 94.9760 & 95.0160 & +0.0401 \\
\midrule
Mean                    & 95.5151 & 95.5117 & -0.0034 \\
\bottomrule
\end{tabular}
\end{table}

The validation results show no broad performance shift in favor of either
model. Instead, small positive and negative differences occur across
individual conditions.

\FloatBarrier
\subsection{Clean-Test Performance}

Table~\ref{tab:clean_results} reports performance on the unmodified official
EMNIST Letters test partition.

\begin{table}[htbp]
\centering
\caption{Clean official-test performance of the matched cross-entropy
baseline and CAML.}
\label{tab:clean_results}
\setlength{\tabcolsep}{7pt}
\renewcommand{\arraystretch}{1.25}
\begin{tabular}{lrr}
\toprule
\textbf{Metric} &
\textbf{Baseline} &
\textbf{CAML} \\
\midrule
Accuracy (\%)      & 95.6346 & \textbf{95.6731} \\
Macro-F1           & 0.956326 & \textbf{0.956724} \\
Number correct     & 19,892 & 19,900 \\
Number incorrect   & 908 & 900 \\
\bottomrule
\end{tabular}
\end{table}

CAML increases clean-test accuracy by \(0.0385\) percentage points, equivalent
to eight additional correct classifications among 20,800 test examples.
Macro-F1 changes by approximately \(4.0\times10^{-4}\).

The paired prediction analysis shows that this small net gain results from a
substantially larger exchange of individual decisions. CAML corrects 124
examples that the baseline classifies incorrectly, but it also changes 116
baseline-correct predictions into errors. Thus,

\[
n_{01}=124,
\qquad
n_{10}=116.
\]

The exact McNemar test gives \(p=0.6515\). The observed clean-test difference
is therefore not statistically significant and should be interpreted as a
small directional change rather than evidence of a reliable global accuracy
improvement.

\subsection{Robustness Under Controlled Perturbations}

Table~\ref{tab:robustness_results} reports performance under the ten predefined
corrupted test conditions.

\begin{table}[htbp]
\centering
\caption{Official-test accuracy under controlled perturbations.
``Gain'' denotes CAML accuracy minus baseline accuracy in percentage points.
The final column reports Holm-adjusted McNemar \(p\)-values.}
\label{tab:robustness_results}
\setlength{\tabcolsep}{3.4pt}
\renewcommand{\arraystretch}{1.18}
\small
\begin{tabular}{lrrrr}
\toprule
\textbf{Condition} &
\textbf{Baseline} &
\textbf{CAML} &
\textbf{Gain (pp)} &
\textbf{Holm \(p\)} \\
\midrule

Rotation \(5^\circ\)
& 95.5240 & 95.6298 & +0.1058 & 1.0000 \\

Rotation \(10^\circ\)
& 95.5144 & 95.5433 & +0.0288 & 1.0000 \\

Translation 1 px
& 95.4327 & 95.5337 & +0.1010 & 1.0000 \\

Translation 2 px
& 95.3269 & 95.3269 &  0.0000 & 1.0000 \\

Noise \(\sigma=0.05\)
& 95.4663 & 95.5529 & +0.0865 & 1.0000 \\

Noise \(\sigma=0.10\)
& 95.3269 & 95.4375 & +0.1106 & 1.0000 \\

Blur \(\sigma=0.6\)
& 95.5817 & 95.5865 & +0.0048 & 1.0000 \\

Blur \(\sigma=1.0\)
& 95.5240 & 95.4856 & -0.0385 & 1.0000 \\

Stroke thickening
& 95.2788 & 95.4183 & +0.1394 & 0.8051 \\

Stroke thinning
& 94.8365 & 94.7885 & -0.0481 & 1.0000 \\

\bottomrule
\end{tabular}
\end{table}

CAML obtains higher accuracy in seven of the ten corruption conditions, ties
the baseline in one, and performs worse in two. The mean corrupted accuracies
are approximately \(95.3812\%\) for the baseline and \(95.4303\%\) for CAML.
Using the underlying unrounded condition accuracies, the mean difference is
approximately \(+0.0490\) percentage points.

The largest positive difference occurs under stroke thickening
(\(+0.1394\) percentage points), followed by Gaussian noise with
\(\sigma=0.10\) (\(+0.1106\)), rotation by \(5^\circ\) (\(+0.1058\)), and
one-pixel translation (\(+0.1010\)). CAML is slightly worse under
\(\sigma=1.0\) blur and stroke thinning.

The direction of most point estimates is positive, but the paired statistical
analysis does not establish a robust advantage. Even the largest unadjusted
difference, under stroke thickening, has \(p=0.0805\). After Holm correction,
its adjusted value is \(p_{\mathrm{Holm}}=0.8051\), and all remaining adjusted
values equal \(1.0\). None of the ten condition-specific comparisons is
therefore statistically significant after controlling for multiple testing.

\subsection{Paired Error Transitions}

Aggregate accuracy alone does not show how often the two models disagree about
individual examples. Table~\ref{tab:paired_transitions} reports paired
correctness transitions for each condition.

\begin{table}[htbp]
\centering
\caption{Paired prediction transitions from the baseline to CAML.
``Fixed'' denotes baseline-wrong/CAML-correct examples; ``Broken'' denotes
baseline-correct/CAML-wrong examples.}
\label{tab:paired_transitions}
\setlength{\tabcolsep}{5pt}
\renewcommand{\arraystretch}{1.18}
\begin{tabular}{lrrr}
\toprule
\textbf{Condition} &
\textbf{Fixed} &
\textbf{Broken} &
\textbf{Net} \\
\midrule
Clean                  & 124 & 116 &  +8 \\
Rotation \(5^\circ\)   & 133 & 111 & +22 \\
Rotation \(10^\circ\)  & 134 & 128 &  +6 \\
Translation 1 px       & 125 & 104 & +21 \\
Translation 2 px       & 141 & 141 &   0 \\
Noise \(\sigma=0.05\)  & 127 & 109 & +18 \\
Noise \(\sigma=0.10\)  & 151 & 128 & +23 \\
Blur \(\sigma=0.6\)    & 118 & 117 &  +1 \\
Blur \(\sigma=1.0\)    & 118 & 126 &  -8 \\
Stroke thickening      & 143 & 114 & +29 \\
Stroke thinning        & 159 & 169 & -10 \\
\bottomrule
\end{tabular}
\end{table}

The auxiliary objective does not merely correct a small fixed subset of
baseline errors. In every condition, more than 200 examples change correctness
status between the two models. For example, on clean data CAML fixes 124
baseline errors but introduces 116 new ones. Under Gaussian noise with
\(\sigma=0.10\), it fixes 151 and breaks 128, whereas under stroke thinning it
fixes 159 and breaks 169.

These paired transitions show that CAML changes individual predictions even
when aggregate accuracy changes very little. This behavior is consistent with
a redistribution of class decisions, but by itself it does not establish a
specific geometric change in the embedding space.

\subsection{Effect on the Discovered Confusion Pairs}
\label{subsec:pair_results}

Because CAML specifically targets the discovered rival relationships, overall
accuracy alone is insufficient for evaluating its intended effect. We
therefore examine the clean-test symmetric error counts for the same eight
pairs selected before final-model training.

Table~\ref{tab:selected_pair_test} retains the discovery ranking used in
Table~\ref{tab:discovered_pairs} to make the two analyses directly comparable.

\begin{table}[htbp]
\centering
\caption{Clean-test symmetric error counts for the eight confusion pairs
identified on the independent discovery partition. A negative change denotes
fewer errors under CAML. Pair-level differences are reported descriptively and
are not presented as separately significant effects.}
\label{tab:selected_pair_test}
\setlength{\tabcolsep}{4.5pt}
\renewcommand{\arraystretch}{1.22}
\begin{tabular}{lrrrr}
\toprule
\textbf{Pair} &
\textbf{Graph weight} &
\textbf{Baseline errors} &
\textbf{CAML errors} &
\textbf{Change} \\
\midrule
I/L & 1.0000 & 344 & 334 & -10 \\
G/Q & 0.3654 & 137 & 145 &  +8 \\
U/V & 0.0865 &  49 &  50 &  +1 \\
R/V & 0.0865 &  11 &   9 &  -2 \\
D/O & 0.0769 &  20 &  17 &  -3 \\
C/E & 0.0673 &   7 &   6 &  -1 \\
I/J & 0.0673 &  25 &  22 &  -3 \\
H/N & 0.0577 &  21 &  22 &  +1 \\
\midrule
\textbf{Total} & -- & \textbf{614} & \textbf{605} & \textbf{-9} \\
\bottomrule
\end{tabular}
\end{table}

Across the selected confusion set, the total symmetric error count decreases
from 614 to 605, corresponding to a relative reduction of

\[
\frac{614-605}{614}\times100
=
1.4658\%.
\]

The pair-level response is heterogeneous. The dominant I/L relationship
improves from 344 to 334 errors, a reduction of approximately \(2.91\%\).
Several smaller pairs also improve: I/J changes from 25 to 22, D/O from 20 to
17, R/V from 11 to 9, and C/E from 7 to 6.

The second-largest discovered relationship, G/Q, moves in the opposite
direction, increasing from 137 to 145 errors. U/V and H/N each increase by one
error.

These results indicate that the targeted objective changes error behavior in
the intended aggregate direction but does not improve every selected edge.
In particular, a larger discovery weight does not guarantee a favorable
downstream response. The pair-specific counts should therefore be interpreted
as descriptive evidence about where the predictions changed, not as
independent statistically established improvements.

\subsection{Training Behavior of the CAML Constraint}

The CAML objective also provides a direct measure of how frequently the
selected pairwise constraints remain unsatisfied during optimization.

At the beginning of training, approximately \(23.1\%\) of active
confusion-pair constraints are violated on an average minibatch. This fraction
declines throughout training and is below \(10\%\) near the selected
checkpoint; by epoch 27, the mean violation fraction is approximately
\(8.95\%\).

The auxiliary margin-loss term also decreases, from approximately \(0.0813\)
during the first epoch to approximately \(0.0505\) by epoch 27.

These training statistics show that the auxiliary objective is actively
optimized rather than remaining numerically inert. They are consistent with
increased satisfaction of the specified cosine-separation constraints.
However, they do not by themselves prove a globally improved embedding
geometry. The reduction in constraint violations accompanies only a
\(0.0385\)-percentage-point clean-accuracy increase and a \(1.47\%\) reduction
in the selected-pair error count.

This distinction is central to the experiment: successful optimization of the
explicit CAML constraint does not necessarily translate into a statistically
reliable improvement in top-1 recognition.

\section{Discussion}
\label{sec:discussion}

\subsection{What CAML Changes}

The results show that CAML is not numerically inactive. During training, the
fraction of violated pair-specific constraints decreases substantially, and
the clean-test error distribution over the selected confusion pairs changes
relative to the matched baseline. These observations show that the auxiliary
objective affects optimization and alters pair-specific predictions.

The effect should not, however, be interpreted as evidence of a broad
improvement in recognition. On clean data, CAML corrects 124 baseline errors
while introducing 116 new errors. Similar exchanges occur under the corrupted
conditions. Consequently, the net changes in top-1 accuracy remain small even
though hundreds of individual predictions differ between the two models.

The most appropriate interpretation is therefore that CAML changes how errors
are distributed around selected class relationships. The observed decline in
margin violations is consistent with increased satisfaction of the prescribed
cosine-separation constraints, but the present experiment does not directly
establish a globally improved embedding geometry.

\subsection{Why Targeted Margins Produce Only a Small Aggregate Gain}

One explanation for the limited aggregate effect is the strength of the
baseline. The matched CNN already achieves \(95.6346\%\) clean accuracy and
maintains approximately \(95.4\%\) mean accuracy across the tested
perturbations. Only a relatively small fraction of examples therefore remains
available for correction.

More importantly, class decisions are coupled. Increasing separation between
one class and a selected rival can also alter the relative position of an
example with respect to other classifier directions. A constraint intended to
improve one relationship may therefore create compensating changes elsewhere.

The contrast between I/L and G/Q illustrates this behavior. I/L is the
strongest edge in the discovery graph and improves from 344 to 334 symmetric
clean-test errors. G/Q is the second-strongest edge, yet its error count
increases from 137 to 145. Thus, confusion frequency alone does not determine
whether a larger pairwise cosine-separation requirement will improve the final
classification boundary.

Residual handwriting errors may arise from several mechanisms, including
within-class variation, intrinsically ambiguous glyphs, missing local evidence,
or interactions among several neighboring classes. A single global embedding
constraint cannot explicitly distinguish among these causes.

\subsection{Pair-Specific Improvement Versus Global Improvement}

The experiment also demonstrates why a confusion-aware method should not be
evaluated only through aggregate accuracy. CAML is designed to act on a sparse
set of selected class relationships, so pair-specific behavior provides useful
descriptive evidence about whether those relationships change.

Five of the eight selected pairs show fewer symmetric clean-test errors under
CAML, and the total over the selected set decreases from 614 to 605. At the
same time, three selected pairs become worse, including G/Q, the
second-strongest discovery edge.

This distinction can be summarized as

\begin{equation}
\text{targeted pair improvement}
\not\Rightarrow
\text{global accuracy improvement}.
\label{eq:pair_vs_global}
\end{equation}

A training objective can improve some intended relationships while producing
offsetting changes elsewhere. In the present experiment, the aggregate
selected-pair reduction is \(1.47\%\), whereas the clean top-1 accuracy change
is only \(+0.0385\) percentage points and is not statistically significant.

The pair-level reductions are reported descriptively. No separate inferential
test is used to claim that an individual pair improvement, or the aggregate
selected-pair reduction, is statistically established.

\subsection{Interpretation of the Robustness Results}

The robustness evaluation shows a similar pattern. CAML has a positive point
estimate in seven of the ten corrupted conditions, ties the baseline in one,
and has a negative point estimate in two. The mean corrupted accuracy is
slightly higher for CAML, but the magnitude of the difference is small.

The largest observed gain occurs under stroke thickening, while smaller
positive differences occur under rotation, translation, and additive Gaussian
noise. Performance decreases slightly under the stronger blur condition and
under stroke thinning.

These results do not support a claim that CAML systematically improves
robustness. None of the ten condition-specific differences remains
statistically significant after Holm correction. Instead, the perturbation
suite shows that the CAML model remains closely matched to the baseline across
the tested transformations and does not obtain its small clean-data gain
through a large degradation in the evaluated corrupted conditions.

\subsection{Why I/L and G/Q Respond Differently}

The contrasting behavior of I/L and G/Q suggests that different confusion
pairs may require different forms of corrective supervision.

For I/L, a substantial part of the ambiguity may be represented by a relatively
direct competition between two classifier directions. Increasing the required
cosine separation can therefore help some examples move toward the correct
side of that competition.

G/Q may involve a more localized distinction. Handwritten instances of the two
letters can share a dominant rounded structure while differing through a small
opening, tail, or stroke configuration. CAML operates on the final
256-dimensional global embedding and does not explicitly identify which image
region contains the evidence needed to distinguish the two classes.

The deterioration of G/Q therefore does not imply that the pair was incorrectly
identified as difficult. Rather, it suggests that identifying a difficult
relationship and identifying the appropriate mechanism for correcting it are
different problems. A large confusion rate signals where the classifier
struggles, but it does not uniquely determine the form or strength of the
representation constraint that should be applied.

\subsection{Relation to Adaptive and Confusion-Aware Margin Learning}

CAML should be interpreted within the broader family of adaptive-margin and
hard-example objectives rather than as the first method to vary supervision
according to class or sample difficulty. Existing approaches adapt margins
according to class statistics, emphasize difficult negatives, modify emphasis
over the course of training, or incorporate confusion information directly
into margin construction
\cite{liu2019adaptiveface,wang2020mvsoftmax,huang2020curricularface,
wang2025cacm,zhao2026confas}.

The distinguishing feature studied here is the combination of three design
choices. First, CAML represents difficulty as a sparse graph of
\emph{specific rival class pairs} rather than as a single class-level or
sample-level difficulty value. Second, the graph is estimated on a dedicated
discovery partition before final-model training. Third, its effect is isolated
through a matched single-network comparison with no additional inference-time
machinery.

This design allows the experiment to ask a narrow question: whether a fixed,
training-side estimate of pairwise confusion can be converted into useful
pair-specific cosine constraints. The present results provide partial support
for that mechanism at the level of targeted error redistribution, but not
evidence of a statistically reliable global performance advantage.

\subsection{Advantages of the CAML Formulation}

Despite the small measured performance difference, the CAML formulation has
several methodological advantages.

First, the targeted relationships are selected automatically from data rather
than manually copied from the final test confusion matrix.

Second, pair discovery is separated from final validation and official test
evaluation. This makes the origin of the confusion graph explicit and reduces
the risk of test-driven pair selection within the CAML experiment.

Third, CAML introduces no additional trainable inference module. The final
baseline and CAML classifiers contain the same number of parameters and use
the same prediction rule.

Fourth, the auxiliary objective is interpretable. Every nonzero CAML term
corresponds to an observed class-pair relationship, and its relative weight is
determined by the discovery confusion rate.

Finally, the formulation makes limited or negative results informative. A
declining CAML loss together with little change in final accuracy indicates
that successful optimization of the specified pairwise constraint is not, by
itself, sufficient to guarantee better recognition.

\section{Limitations}
\label{sec:limitations}

Several limitations should be considered when interpreting the experiment.

\subsection{Single Dataset}

The study is conducted only on EMNIST Letters. The dataset is appropriate for
investigating visually similar character classes, but the results cannot be
assumed to generalize to other handwriting datasets, natural-image
classification, or tasks with substantially different class structures.

A stronger test of the general CAML principle would apply the same discovery
and training protocol to additional multiclass datasets without modifying the
method in response to their final test outcomes.

\subsection{Single Random Seed}

The reported experiment uses one fixed random seed. Although the baseline and
CAML models are tightly matched within the run, neural-network optimization is
stochastic. A clean-test difference of only a few hundredths of a percentage
point may be smaller than ordinary variation across independently initialized
training runs.

The observed \(+0.0385\)-percentage-point difference should therefore not be
interpreted as an estimate of a stable expected CAML advantage. A multi-seed
study would be required to estimate the distribution of the CAML--baseline
difference across independent runs.

The McNemar tests condition on the particular trained baseline--CAML model
pair; they quantify paired test-set prediction differences but do not account
for variability across independent training runs.

\subsection{Fixed Confusion-Graph Size}

The graph is restricted to the eight highest-ranked eligible confusion pairs.
This produces a sparse and interpretable experiment, but the value eight is a
design choice rather than a theoretically optimal graph size.

A smaller graph would focus the auxiliary objective more strongly on the
largest relationships, whereas a larger graph would distribute margin pressure
across more class pairs. The present study does not perform a post-hoc graph
size sweep because selecting the graph size after repeatedly inspecting the
official test results would weaken the independence of the final evaluation.

\subsection{Fixed Margin Hyperparameters}

The experiment uses

\[
m_0=0.10,
\qquad
m_1=0.10,
\qquad
\lambda=0.20.
\]

These values determine the base cosine margin, the confusion-dependent margin
increment, and the weight of the auxiliary objective, respectively. They are
not claimed to be globally optimal.

A systematic hyperparameter search would require an appropriate development
procedure that preserves final-test independence. The present results should
therefore be interpreted as the behavior of one predefined CAML
configuration rather than the maximum performance obtainable through margin
tuning.

\subsection{Discovery-Set Sampling Variability}

The confusion graph is estimated from 6,240 discovery examples. Although this
partition is stratified and independent of pilot training, several weaker
selected edges are supported by fewer than ten symmetric discovery errors.

These low-count relationships may be sensitive to the particular discovery
sample. Repeating graph construction across independent training-side splits
would allow the stability of both edge selection and edge weighting to be
quantified.

\subsection{Global Rather Than Local Constraints}

CAML acts on the final 256-dimensional image embedding. Its auxiliary
constraint therefore operates at the level of a global representation and does
not identify which spatial region of a character is responsible for a
particular confusion.

This limitation may be important when two letters differ through only a small
local component. Future work could combine pair discovery with localized
attention, part-specific representations, or spatial contrastive constraints.
Those extensions are outside the scope of the present experiment.

\subsection{No Pair-Level Inferential Testing}

The pair-specific clean-test counts are used to examine whether the selected
relationships change in the intended direction. These comparisons are
descriptive. The present study does not perform separate hypothesis tests over
the eight selected pairs.

Consequently, reductions such as I/L from 344 to 334 errors, or the aggregate
selected-pair reduction from 614 to 605, should not be interpreted as
independently established statistically significant improvements.

\section{Threats to Validity}
\label{sec:validity}

The principal internal-validity concern for a confusion-aware method is
evaluation leakage into pair selection. Within the CAML protocol, this concern
is addressed by assigning distinct roles to the core, discovery, validation,
and official test partitions.

The confusion graph is constructed only from predictions of the pilot model on
the discovery subset. The final validation set does not determine the graph,
and the official test set is not used within the CAML experiment to select
confusion pairs, checkpoints, or margin weights.

A second concern is unequal training conditions between the proposed method
and its baseline. To reduce this confound, the final baseline and CAML models
use the same architecture, parameter count, final training examples, minibatch
ordering, deterministic augmentation schedule, optimizer, learning rate,
weight decay, label smoothing, learning-rate scheduler, validation criterion,
and early-stopping rule. Their intended experimental difference is the
presence or absence of the CAML auxiliary objective.

A third concern is the interpretation of very small numerical differences.
The clean-test advantage corresponds to only eight additional correct
predictions among 20,800 examples and is not significant under the paired
McNemar test. The robustness differences are similarly small and do not remain
significant after Holm correction. The manuscript therefore treats these
values as point estimates rather than evidence of reliable superiority.

External validity is more limited. Only one dataset, one CNN architecture, one
discovery split, and one random seed are evaluated. The findings should
therefore be interpreted as evidence about this CAML configuration under the
stated EMNIST Letters protocol, not as a universal claim about
confusion-adaptive margin learning.

\section{Reproducibility}
\label{sec:reproducibility}

The experiment is implemented in PyTorch using a fixed random seed of 42. The
recorded outputs include the data split, pilot confusion matrix, selected
graph edges and weights, training histories, selected checkpoints, robustness
results, paired prediction transitions, and clean-test selected-pair analysis.

Table~\ref{tab:reproducibility} summarizes the principal configuration.

\begin{table}[H]
\centering
\caption{Principal reproducibility settings for the CAML experiment.}
\label{tab:reproducibility}
\setlength{\tabcolsep}{6pt}
\renewcommand{\arraystretch}{1.22}
\begin{tabular}{ll}
\toprule
\textbf{Setting} & \textbf{Value} \\
\midrule
Random seed & 42 \\
Dataset & EMNIST Letters \\
Official training size & 124,800 \\
Official test size & 20,800 \\
Core size & 106,080 \\
Discovery size & 6,240 \\
Validation size & 12,480 \\
Final training size & 112,320 \\
Pilot epochs & 5 \\
Maximum final epochs & 30 \\
Batch size & 256 \\
Evaluation batch size & 512 \\
Optimizer & AdamW \\
Initial learning rate & \(10^{-3}\) \\
Weight decay & \(10^{-4}\) \\
Label smoothing & 0.05 \\
Early-stopping patience & 6 \\
Selected confusion pairs & 8 \\
Base cosine margin \(m_0\) & 0.10 \\
Adaptive increment \(m_1\) & 0.10 \\
CAML weight \(\lambda\) & 0.20 \\
Embedding dimension & 256 \\
Trainable parameters & 1,908,730 \\
\bottomrule
\end{tabular}
\end{table}

The final experimental artifacts include the baseline and CAML checkpoints,
training histories, discovered confusion graph, validation summaries,
clean-test predictions, selected-pair analysis, robustness comparisons, and
machine-readable summary statistics.

The public executable notebook used to reproduce the CAML experiment is
provided in the Code Availability section at the end of the manuscript.

\FloatBarrier

\section{Conclusion}
\label{sec:conclusion}

This study introduced \textbf{Confusion-Adaptive Margin Learning (CAML)}, a
training strategy that converts empirically discovered class-pair confusion
into targeted representation-space constraints. The method extends our earlier
work on curvature--orientation representations for handwritten-character
recognition \cite{nouri2025ijca,NouriCurv2025} by shifting attention from
explicit feature construction to the structure of the learning objective.

CAML uses a dedicated discovery partition, held out from pilot training, to
construct a sparse weighted graph of difficult class relationships. The graph
is fixed before final-model training and is converted into pair-specific
cosine-separation constraints. The final classifier requires no specialist
models, routing rules, or additional inference-time parameters.

Under the controlled EMNIST Letters protocol, CAML slightly increases clean
accuracy from \(95.6346\%\) to \(95.6731\%\) and reduces the aggregate
symmetric error count over the eight targeted pairs from 614 to 605. The
largest discovered pair, I/L, improves, whereas the second-largest pair, G/Q,
deteriorates. Across the ten corrupted test conditions, CAML produces positive
accuracy differences more often than negative ones, but none of the
condition-specific differences remains statistically significant after Holm
correction. The clean paired comparison is likewise not significant.

The principal result is therefore not that CAML establishes a superior
handwriting recognizer. Rather, the experiment demonstrates that a
confusion-derived pair-specific objective can alter targeted prediction
behavior while producing only limited aggregate benefit when applied to an
already strong classifier. Successful satisfaction of a margin constraint is
not equivalent to statistically reliable improvement in recognition accuracy.

This distinction suggests an important direction for future confusion-aware
learning. Confusion statistics identify \emph{where} a classifier struggles,
but they do not necessarily identify \emph{why} the confusion occurs or which
representation change will correct it. Future work should therefore examine
the stability of confusion graphs across independent seeds and datasets,
investigate interactions among multiple neighboring classes, and consider
localized or part-specific constraints for class distinctions that depend on
small spatial features.
\section*{Code Availability}

The executable notebook used for the CAML experiment is publicly available in
the project repository:

\begin{center}
\url{https://github.com/Meetra21/Curvature-Orientation-MLP}
\end{center}

The experiment is contained in
\texttt{CAML\_EMNIST\_Confusion\_Adaptive\_Margin.ipynb}.
See related articles and supporting materials in \cite{NouriSobel2025}
\cite{nouri2025critical}
\cite{nouri2026siggain}
\cite{nouri2026mdl}
\cite{nouri2026stepcache}.

\end{document}